\documentclass{IOS-Book-Article}

\usepackage{mathptmx}
\usepackage{soul}\setuldepth{article}
\usepackage{hyperref}
\usepackage{url}
\usepackage{amsmath}
\usepackage{xcolor}
\usepackage{graphicx}
\usepackage{amssymb}
\usepackage{comment}
\usepackage{makecell}
\newtheorem{example}{Example}
\newtheorem{definition}{Definition}

\usepackage{comment}

\def\hb{\hbox to 11.5 cm{}}

\newcounter{todocounter}
\newcounter{genetcounter}
\newcommand{\ma}[1]{\stepcounter{todocounter}
  {\color{blue} Mehwish: \thetodocounter: #1}}
\newcommand{\ph}[1]{\textcolor{red}{PH: #1}}

\begin{document}

\pagestyle{headings}
\def\thepage{}
\begin{frontmatter}              

\title{Neurosymbolic Methods for Dynamic Knowledge Graphs}


\markboth{}{July 2024\hb}

\author[A]{\fnms{Mehwish} \snm{Alam} 
\thanks{Corresponding Author: mehwish.alam@telecom-paris.fr.}}
,
\author[B]{\fnms{Genet Asefa} \snm{Gesese} 
}
and
\author[C]{\fnms{Pierre-Henri} \snm{Paris} 
}


\runningauthor{M. Alam et al.}
\address[A]{Télécom Paris, Institut Polytechnique de Paris, France}
\address[B]{FIZ Karlsruhe, Karlsruhe Institute of Technology, Germany}
\address[C]{University of Paris Saclay, France}

\begin{abstract}
Knowledge graphs (KGs) have recently been used for many tools and applications, making them rich resources in structured format. However, in the real world, KGs grow due to the additions of new knowledge in the form of entities and relations, making these KGs dynamic. This chapter formally defines several types of dynamic KGs and summarizes how these KGs can be represented. Additionally, many neurosymbolic methods have been proposed for learning representations over static KGs for several tasks such as KG completion and entity alignment. This chapter further focuses on neurosymbolic methods for dynamic KGs with or without temporal information. More specifically, it provides an insight into neurosymbolic methods for dynamic (temporal or non-temporal) KG completion and entity alignment tasks. It further discusses the challenges of current approaches and provides some future directions. 
\end{abstract}

\begin{keyword}
Dynamic Knowledge Graphs\sep Temporal Knowledge Graphs\sep Dynamic Entity Alignment \sep Dynamic Knowledge Graph Completion
\end{keyword}
\end{frontmatter}
\markboth{July 2024\hb}{July 2024\hb}

\section{Introduction}\label{sec:introduction}

Knowledge Graphs (KGs)~\cite{csur/HoganBCdMGKGNNN21} have gained attention in the past few years for representing information in a structured way, i.e., entities and relations. It has been used for various applications and domains from search engines and recommendation systems~\cite{semweb/IanaAP24} to bio-informatics~\cite{ws/JonquetLFCNMS11} and social sciences~\cite{snam/ChenSA22}. They capture relationships between entities in a way that is both human-readable and machine-processable, enabling advanced reasoning and inference capabilities. The significance of KGs lies in their ability to integrate vast amounts of heterogeneous data, providing a unified framework that supports comprehensive querying and analysis.

However, the real world is dynamic, with changes happening continuously—new entities emerge, existing relationships evolve, and facts that were once true may become outdated. In such scenarios, traditional static KGs fall short as they cannot accommodate these changes in real-time. This limitation has led to the development of \textit{Dynamic Knowledge Graphs (DKGs)}, which can evolve by incorporating temporal information. Dynamic KGs update the data and track changes, enabling temporal queries and historical analysis.

The inclusion of temporality in KGs poses significant challenges. Firstly, the representation of time-sensitive data requires precise modeling to capture the validity period of facts. For example, without temporal information, a query asking about the president of the United States might return multiple results, leading to ambiguity. Temporal dynamics are crucial in contexts like legal documents, medical records, and financial transactions, where the timing of events plays a pivotal role in decision-making. Thus, preserving temporality in KGs is beneficial and necessary for ensuring the accuracy and relevance of the information.

Furthermore, learning representations over DKGs is essential for several reasons. It enables the development of models that can predict future changes, discover hidden patterns, and provide insights based on the temporal evolution of data. This is particularly valuable in predictive analytics, trend analysis, and anomaly detection, where understanding the temporal context can significantly enhance the accuracy of the results.

Entity alignment, the task of identifying equivalent entities across different datasets or KGs, becomes increasingly complex in dynamic settings. As entities and their attributes change over time, ensuring they are correctly matched requires sophisticated techniques that can handle temporal variations. This is crucial for maintaining data integrity, improving the accuracy of merged datasets, and enhancing the overall quality of the KG.

In summary, the dynamic and temporal dimensions of KGs are critical for accurately reflecting the complexity of real-world data. Addressing the challenges associated with these aspects can significantly advance the field of KG research and its applications, paving the way for more intelligent and responsive information systems.

This book chapter is structured as follows: Section~\ref{sec:related_work} reviews related work and positions this chapter within the broader context. Section~\ref{sec:prelimiaries} provides fundamental definitions of the DKGs. Section~\ref{sec:representation} details how the dynamic information can be represented in the form of a KG. Following the discussion on representation techniques, the book chapter then moves on to explaining how neurosymbolic methods have been utilized for learning representations over DKGs for KG Completion (Section~\ref{sec:kgcompletion}) and dynamic entity alignment (Section~\ref{sec:entity_alignment}).

\section{Related Work}\label{sec:related_work}

One of the very recent studies provides an overview of the DKGs~\cite{tgdk/PolleresPBDDDEF23}. However, this chapter mostly focuses on providing a formal definition of different categories of DKGs along with neurosymbolic methods for DKGs, such as learning representations over such symbolic representations using methods based on neural networks along with the downstream tasks such as KG completion and entity alignment. 

Several studies have been conducted on learning representations over KGs. For instance, various techniques for refining KGs are summarized in~\cite{Paulheim17}, including methods for KG completion. 
A categorization of static KG completion algorithms is given in~\cite{WangMWG17}. In contrast, an exhaustive survey on multimodal KG embedding algorithms that utilize literal (text, numeric, image) information within the static KGs are presented in \cite{GeseseBAS21} along with an experimental comparison of these algorithms. Furthermore, a recent article~\cite{corr/abs-2308-00081} provides an overview of various KG completion tasks, such as transductive and inductive link prediction. It details methods incorporating background information from large language models and discusses embeddings considering description logic axioms. However, all these studies are limited to static KGs and do not consider the dynamic aspects in a KG. In contrast,~\cite{Cai0GZLL23} surveys the methods for Temporal Knowledge Graph (TKG) completion. This book chapter builds on this by formally defining the fundamental differences between DKGs and TKGs. It provides a comprehensive overview of representation learning for these types of KGs, highlighting their unique characteristics and challenges. Additionally, the chapter discusses various downstream tasks associated with these graphs, such as KG completion and entity alignment, offering insights into the methodologies and applications tailored to DKGs and TKGs.


\section{Preliminaries}\label{sec:prelimiaries}
This section formally defines static, temporal, and dynamic KGs with examples.

\begin{definition}[Static Knowledge Graph]
\label{def:static-kg}

Let $G=(E,R,L,F)$ be a directed labeled graph, where $E$, $R$, and $L$ are the sets of entities, relations, and literals, respectively. $ F \subseteq E \times R \times (E \cup L)$ represents a set of facts such that $f =(h,r,t)$ or $f =(h,r,l)$ represent one triple where $f \in F$, $h,t \in E $, $r \in R$, and $l \in L$. 

\end{definition}

\begin{example}\label{ex:static_kg}
    Consider a simple knowledge graph $G = (E, R, L, F)$ where:
    \begin{itemize}
        \item \textbf{Entities (E)}: \{Barack\_Obama, USA\}
        \item \textbf{Relations (R)}: \{president\_of\}
        \item \textbf{Literals (L)}: $\emptyset$
        \item \textbf{Facts (F)}:  \{(Barack\_Obama, president\_of, USA)\}
    \end{itemize}
    The example shows that Barack Obama is the president of the USA.
    Now, suppose we update the graph with the following:
    \begin{itemize}
        \item $E=E\cup\{Donald\_Trump\}$
        \item $R=R$
        \item $L=\emptyset$
        \item $F=F\cup(Donald\_Trump, president\_of, USA)$
    \end{itemize}
    In this static representation, the knowledge graph indicates that both Barack Obama and Donald Trump have been (or are) presidents of the USA. 
    However, without temporal information, it is unclear when each individual held office.
\end{example}


The definition of a static KG can further be extended to a temporal KG, where each triple can have a time interval representing its temporal validity:


\begin{definition}[Temporal Knowledge Graph]
\label{def:temporal-kg}
Let $G = (E,R,L,T,Q)$ be a directed labeled graph, where $E$, $R$, $L$, and $T$ represent the set of entities, relations, literals, and timestamps, respectively. $ Q \subseteq E \times R \times (E\cup L) \times (T\cup\{\emptyset\})$ represents a set of facts such that $q = (h,r,t,[\tau_s,\tau_e ])$ or $q = (h,r,l,[\tau_s,\tau_e ])$ represent one quadruple where $q\in Q$, $h,t \in E $, $r \in R$, $l\in L$, and $\tau_s, \tau_e \in T\cup \{\emptyset \}$ and $\tau_s$ and $\tau_e$ represent the start and end time, i.e., $\tau_s \leq \tau_e$ . If $\tau_s = \tau_e$ then it represents a point in time $\tau \in T$.  
\end{definition}


\begin{example}\label{ex:temporal_kg}
    Consider a simple knowledge graph \( G = (E, R, L, F) \) where:

    \begin{itemize}
        \item \textbf{Entities (E)}: \{Barack\_Obama, USA\}
        \item \textbf{Relations (R)}: \{president\_of\}
        \item \textbf{Literals (L)}: \{2009, 2017\} (representing years)
        \item \textbf{Quadruples (Q)}: \{(Barack\_Obama, president\_of, USA, [2009, 2017])\}
    \end{itemize}
    The example shows when the fact that Barack Obama held office is a valid fact, i.e.,~from 2009 to 2017.
\end{example}

\begin{definition}[Dynamic Knowledge Graph]
    A dynamic knowledge graph $G$ is a sequence of knowledge graphs $\{G_{t_0}, G_{t_1}, \ldots, G_{t_n}\}$ indexed by time steps $t_0, t_1, \ldots, t_n$ such that $t_0 < t_1 < \cdots < t_n$ and $\forall t_i, t_i\in T$, where $T$ is the set of all timestamps. 
    Each graph $G_{t_i}$, at time step $t_i$, can be either a static knowledge graph (see Definition~\ref{def:static-kg}) or a temporal knowledge graph (see Definition~\ref{def:temporal-kg}).
    
    The graph $G_{t_i}$ represents a snapshot of the dynamic graph $G$ at time $t_i$. The transition between two consecutive snapshots $G_{t_{i-1}}$ and $G_{t_i}$ can be characterized by changes in entities, relations, literals, and facts. 
    Formally:
    \begin{align*}
        E_{t_i} & = E_{t_{i-1}} \cup E^{\text{+}}_{t_i} \setminus E^{\text{-}}_{t_{i-1}} \quad \text{where } E^{\text{+}}_{t_i} \text{ and } E^{\text{-}}_{t_{i-1}} \text{ represent the additions and removals of entities}, \\
        R_{t_i} & = R_{t_{i-1}} \cup R^{\text{+}}_{t_i} \setminus R^{\text{-}}_{t_{i-1}} \quad \text{where } R^{\text{+}}_{t_i} \text{ and } R^{\text{-}}_{t_{i-1}} \text{ represent the additions and removals of relations}, \\
        L_{t_i} & = L_{t_{i-1}} \cup L^{\text{+}}_{t_i} \setminus L^{\text{-}}_{t_{i-1}} \quad \text{where } L^{\text{+}}_{t_i} \text{ and } L^{\text{-}}_{t_{i-1}} \text{ represent the additions and removals of literals}, \\
        G_{t_i} & = G_{t_{i-1}} \cup G^{\text{+}}_{t_i} \setminus G^{\text{-}}_{t_{i-1}} \quad \text{where } G^{\text{+}}_{t_i} \text{ and } G^{\text{-}}_{t_{i-1}} \text{ represent the additions and removals of facts}.
    \end{align*}
    
    Additionally, if $G_{t_i}$ is a temporal knowledge graph, it can include changes in the temporal intervals of the quadruples.
\end{definition}

\begin{example}
    Consider the first snapshot of a simple knowledge graph $G_0 = (E_0, R_0, L_0, F_0)$ where:
    \begin{itemize}
        \item \textbf{Entities ($E_0$)}: \{Barack\_Obama, USA\}
        \item \textbf{Relations ($R_0$)}: \{president\_of\}
        \item \textbf{Literals ($L_0$)}: $\emptyset$
        \item \textbf{Facts ($F_0$)}:  \{(Barack\_Obama, president\_of, USA)\}
    \end{itemize}
    We can see that Barack Obama is the president of the USA as in the first example.
    Now, suppose we update the graph with the following new snapshot $G_1 = (E_1, R_1, L_1, F_1)$:
    \begin{itemize}
        \item \textbf{Entities ($E_1$)}: $E_0\setminus\{Barack\_Obama\}\cup\{Donald\_Trump\}$
        \item \textbf{Relations ($R_1$)}: $R_0$
        \item \textbf{Literals ($L_1$)}: $L_0=\emptyset$
        \item \textbf{Facts ($F_1$)}:  \{(Donald\_Trump, president\_of, USA)\}
    \end{itemize}
    Now, Donald Trump is the president of the USA in the last snapshot $G_1$, which no longer contains information about Barack Obama.

    Finally, consider the temporal KG of Example~\ref{ex:temporal_kg} as the first snapshot $G_0$ of a simple knowledge graph.
    It can be updated with the following new snapshot $G_1 = (E_1, R_1, L_1, F_1)$:
    \begin{itemize}
        \item \textbf{Entities ($E_1$)}: $E_0\cup\{Donald\_Trump\}$
        \item \textbf{Relations ($R_1$)}: $R_0$
        \item \textbf{Literals ($L_1$)}: $L_0\cup\{2021\}$
        \item \textbf{Quadruples ($Q_1$)}:  $Q_0\cup\{(Donald\_Trump, president\_of, USA,[2017, 2021])\}$
    \end{itemize}
    This example is more complete because it keeps both facts with their respective time validity.
\end{example}

\section{Representing Dynamic and Temporal Knowledge Graphs}\label{sec:representation}

This section presents the different techniques to represent temporal information in knowledge graphs and that can be used for temporal KGs or dynamic KGs.

\subsection{Techniques to Represent Temporal Information in Knowledge Graphs}

The first nucleus to represent time is the use of various XML Schema Definition~\cite{thompson2004xml} (XSD) datatypes, such as \texttt{xsd:dateTime}, \texttt{xsd:duration}, \texttt{xsd:date}, and others, that allows for precise representation of temporal data, including specific moments, durations, calendar dates, and individual time components, in KGs. These datatypes have been designed to facilitate the accurate and standardized encoding of temporal information, enabling robust temporal querying and reasoning.

\subsubsection{Temporal Properties}

Temporal properties directly incorporate time into relationships within a KG. These properties provide inherent temporal aspects to the entities and relationships they describe. For example (please note that all examples are provided in the Turtle syntax\footnote{\url{https://www.w3.org/TR/rdf12-turtle/}}):
\begin{verbatim}
:Alice :birthDate "1990-01-01"^^xsd:date .
:Alice :employedAt :CompanyX .
:Alice :employmentStart "2020-01-01"^^xsd:date .
:Alice :employmentEnd "2022-01-01"^^xsd:date .
\end{verbatim}

In this example, a simple temporal relation is represented using a time interval:
Alice's employment relationship with CompanyX includes specific start and end dates, enabling queries about her employment duration.
The advantage of this approach is its straightforwardness and direct annotation of temporal data within the relationships themselves.
However, it may lack flexibility for more complex temporal scenarios because one need to know that ``\textit{employmentStart}'' is related to ``\textit{employedAt}''.
Additionally, if one wants to keep track of Alice's employment history, this became more complex to represent.
Temporal attributes can also be represented, e.g.,~we also know her birth date, which is a fixed point in time.

All cross-domain general-purpose KGs have some form of temporal information, e.g.,~Wikidata~\cite{DBLP:conf/www/Vrandecic12}, DBpedia~\cite{DBLP:journals/semweb/LehmannIJJKMHMK15}, YAGO~\cite{DBLP:journals/ai/HoffartSBW13,yago45}, etc.

\subsubsection{Reification}

Reification allows a triple to be treated as a subject of another triple, thereby attaching metadata such as temporal information.
There are several ways to reify a triple.

The standard reification~\cite{DBLP:books/sp/staabS2004/McBride04} is a built-in functionality in RDF, i.e.,~a resource is used to denote the fact
and additional information about the statement can be added using the RDF vocabulary (the properties ``\textit{rdf:subject}'', ``\textit{rdf:predicate}'', and ``\textit{rdf:object}'').
For instance:

\begin{verbatim}
_:statement rdf:type rdf:Statement .
_:statement rdf:subject :Alice .
_:statement rdf:predicate :knows .
_:statement rdf:object :Bob .
_:statement :since "2022-01-01"^^xsd:date .
\end{verbatim}

Through reification, the relationship between Alice and Bob can be annotated with the date they became acquainted.
This approach offers flexibility in adding metadata to existing triples but can lead to increased data redundancy and complexity.

Other possibilities to reify a fact are to use n-ary relations~\cite{DBLP:conf/semweb/ErxlebenGKMV14} or to use singleton properties~\cite{DBLP:conf/www/NguyenBS14}.
The interested reader can refer to Hernández et al.~\cite{DBLP:conf/semweb/HernandezHK15} for more details and an in depth comparison of these approaches.

\subsubsection{Time Ontology in OWL}

The Time Ontology in OWL\footnote{\url{https://www.w3.org/TR/owl-time/}} provides a comprehensive framework for representing temporal concepts.
This ontology includes classes and properties for describing instants, intervals, and durations.
Thus, the ontology is designed to support complex temporal reasoning and querying following Allen's interval algebra~\cite{DBLP:journals/logcom/AllenF94}.
For example:

\begin{verbatim}
@prefix ex: <http://example.org/ns#> .
@prefix time: <http://www.w3.org/2006/time#> .

# Define Alice as a person
ex:Alice a ex:Person ;
    ex:worksSince ex:JobStart .

# Define the job start time
ex:JobStart a time:Instant ;
    time:inXSDDateTime "2020-01-01T00:00:00Z"^^xsd:dateTime .

# Define a class for persons who have been working for more 
# than 3 years
ex:LongTermEmployee a owl:Class ;
    owl:equivalentClass [
        owl:intersectionOf (
            ex:Person
            [
                owl:onProperty ex:worksSince ;
                owl:someValuesFrom [
                    owl:restriction [
                        owl:onProperty time:before ;
                        owl:hasValue 
                            "2021-01-01T00:00:00Z"^^xsd:dateTime
                    ]
                ]
            ]
        )
    ] .

# Reasoning
ex:Alice a ex:LongTermEmployee .  
\end{verbatim}

In this example, we use the OWL Time ontology to describe the start date of Alice's employment and perform basic reasoning to determine if she qualifies as a long-term employee.

The interesting definition is the class ex:LongTermEmployee used to represent individuals who have been working for more than three years.
This class is specified using owl:equivalentClass to intersect ex:Person and a restriction on the ex:worksSince property.
The restriction requires that the ex:worksSince property has a value indicating a work start date before January 1, 2021.

The reasoner determines that since Alice's work start date is ``2020-01-01," which is before the cutoff date ``2021-01-01," she meets the criteria for long-term employment.

\subsubsection{Named Graphs and Quadruples}

Named graphs~\cite{DBLP:journals/ws/CarrollBHS05} group triples into a single graph that can have metadata associated with it.
This allows for temporal information to be added at the graph level.
For example:

\begin{verbatim}
GRAPH <http://example.org/graph/2022-01-01> {
    :Alice :knows :Bob .
}
\end{verbatim}

This structure facilitates temporal data management by associating entire sets of triples with specific temporal contexts.
Named graphs are advantageous for segmenting data into manageable subgraphs but can introduce overhead in graph management and querying.

Quadruples\footnote{\url{https://www.w3.org/TR/rdf12-n-quads/}} extend RDF triples by adding a fourth element, often used to represent the context or the named graph, which can include temporal information.
Contrary to named graphs, quadruples are not a separate graph but a part of the RDF data model, thus, informations can be attached to the fact level instead of the whole graph.
For example:

\begin{verbatim}
:Alice :knows :Bob "2022-01-01"^^xsd:date .
\end{verbatim}

This quadruple includes a timestamp indicating when Alice and Bob's acquaintance began.
The use of quadruples simplifies the representation of contextual information but requires a data store that supports quad storage and querying.

\subsubsection{RDF-star}

RDF-star\footnote{\url{https://w3c.github.io/rdf-star/cg-spec/editors_draft.html}} is an extension of the RDF data model that offers a more compact and intuitive way to represent complex statements, including those involving temporal information.
Examples include:

\begin{verbatim}
<< :Alice :knows :Bob >> :since "2022-01-01"^^xsd:date .
<< :Alice :worksAt :CompanyX >> :startDate "2020-01-01"^^xsd:date ; 
        :endDate "2022-01-01" .
<< :Alice :attended :ConferenceX >> :onDate "2023-06-15"^^xsd:date .
<< :Document123 :hasVersion :Version1 >> :timestamp 
        "2024-01-01T10:00:00Z" .
<< :DataItem :createdBy :User123 >> :creationTime 
        "2024-07-01T09:00:00Z" .
\end{verbatim}

RDF-star enables a streamlined representation of temporal information, simplifying the construction and querying of knowledge graphs.
Its compactness reduces data redundancy and complexity but requires support for RDF-star syntax and semantics in RDF processors.

\subsubsection{Versioning}
Maintaining historical data with timestamps is crucial for tracking changes over time.
Versioning techniques ensure that the evolution of data is documented, enabling temporal queries and historical analysis.
One can use simple approaches, such as adding a timestamp to each triple, or more complex methods, such as reification or named graphs.
For example, using versioning:

\begin{verbatim}
:Document123 :hasVersion :Version1 .
:Version1 :timestamp "2024-01-01T10:00:00Z" .
\end{verbatim}

More complex versioning systems have been proposed over time:
Frommhold et al.~\cite{frommholdVersioningArbitraryRDF2016} developed a Version Control System for KGs, emphasizing the need for efficient change detection across graphs, including handling blank nodes.
Their system uses invertible patches to manage changes, supporting operations like revert and merge,
and ensuring data integrity through secure hashing of patches.
This approach is foundational, particularly in tracking changes in complex datasets and ensuring reliable data history.

Pelgrin et al.~\cite{pelgrinFullyfledgedArchivingRDF2021a} provided a comprehensive survey and analysis of KG archiving systems,
highlighting the lack of standardization and scalability in existing solutions.
They introduced RDFev, a framework for studying the dynamicity of RDF data, which offers insights into dataset evolution at both low and high levels.
Their work outlines the essential features of a fully-fledged RDF archiving system, emphasizing the need for robust support for concurrent updates,
efficient query processing and comprehensive version control features like branching and tagging.

\begin{table}[]
    \resizebox{\textwidth}{!}{%
        \begin{tabular}{lcccccc}
                     & \multicolumn{1}{l}{Properties} & \multicolumn{1}{l}{Reification} & \multicolumn{1}{l}{\begin{tabular}[c]{@{}l@{}}Time\\ Ontology\end{tabular}} & \multicolumn{1}{l}{\begin{tabular}[c]{@{}l@{}}Named Graphs\\ \& Quadruples\end{tabular}} & \multicolumn{1}{l}{RDF-star} & \multicolumn{1}{l}{Versioning} \\ \hline
            Temporal & X                              & X                               & X                                                                           & X*                                                                                       & X                            &                                \\ \hline
            Dynamic  &                                & X                               &                                                                             & X*                                                                                       & X                            & X
        \end{tabular}%
    }
    \caption{Comparison of Temporal and Dynamic Capabilities Across Various RDF Techniques.
        The presence of a feature is indicated by an `X'.
        The asterisk under Named Graphs \& Quadruples denotes that a choice must be made between temporal and dynamic, depending on the semantics intended by the graph owner.\label{tab:representation}}
\end{table}

\vspace{0.5cm}

Table~\ref{tab:representation} provides a summary of the temporal and dynamic capabilities of the various RDF techniques presented in this section
For instance, temporal properties give a straightforward way to annotate data with time, 
but they may not offer the flexibility needed for complex scenarios where relationships between different temporal attributes must be explicitly defined. 
Reification and Named Graphs/Quadruples allow for more detailed metadata, such as temporal context, versioning, or provenance, which can be crucial for historical analyses or data integrity. 
However, these methods can introduce data redundancy and increase the complexity of queries.
Additionally, only one value can be used as a fourth value or as the name of the named graph.

The Time Ontology in OWL is particularly useful for those needing to model complex temporal relations, such as intervals and durations, 
and supports reasoning over these data types. 
This makes it ideal for applications requiring detailed temporal reasoning, such as historical data analysis or event tracking.

RDF-star offers a more compact and intuitive representation, which reduces data redundancy and simplifies the data structure. 
This can be particularly advantageous when dealing with large datasets or when the overhead of traditional reification methods becomes a concern. 
However, RDF processors must support RDF-star syntax and semantics, which might not be universally available.

Versioning techniques are essential for tracking changes over time, ensuring that a KG can evolve while maintaining a record of past states. 
This is particularly relevant in domains where data history is as important as the data itself, such as legal or historical research, or dynamic KG completion.

When choosing a method for representing temporal information in KGs, researchers and practitioners should consider 
the specific requirements of their application, including the complexity of the temporal relationships, the need for metadata, 
the size of the dataset, and the capabilities of the RDF processors they are using.

\subsection{Prominent Knowledge Graphs}

We now describe some of the most prominent general-purpose KGs and how they represent temporal information.

\subsubsection{DBpedia}

DBpedia~\cite{DBLP:journals/semweb/LehmannIJJKMHMK15} incorporates temporal aspects into its dataset through properties that denote specific time-related data.
For example, temporal properties such as \textit{dbo:birthDate} and \textit{dbo:deathDate} indicate birth and death dates for people,
while other properties capture periods of activity or events,
such as \textit{dbo:productionStartDate} and \textit{dbo:productionEndDate }for manufacturing or production events.
This inclusion of temporal data allows for queries about the duration of events or the temporal relationships between entities.
However, representing complex temporal scenarios might require additional context or metadata,
as basic temporal properties might not inherently indicate their relation to other properties without such context.

DBpedia handles versioning primarily by maintaining historical data, which allows tracking changes over time.
The DBpedia Live system plays a crucial role in this aspect, as it continuously synchronizes with Wikipedia to update the dataset with the latest information.
This system utilizes a stream of updates from Wikipedia, enabling DBpedia to reflect recent edits with minimal delay, typically within a few minutes.
Versioning also exists in DBpedia through snapshots, ensuring that historical changes are documented, which is essential for conducting temporal queries and historical analyses.

\subsubsection{YAGO}

The second iteration of YAGO~\cite{DBLP:journals/ai/HoffartSBW13} incorporates temporal dimensions into its KG by assigning existence times to entities and facts.
This is achieved using specific relations such as \textit{wasBornOnDate}, \textit{diedOnDate}, \textit{wasCreatedOnDate}, and \textit{wasDestroyedOnDate},
which are standardized under generic entity-time relations like \textit{startsExistingOnDate} and \textit{endsExistingOnDate}.
For events lasting a single day, YAGO2 uses \textit{happenedOnDate}, which is a sub-property of both \textit{startsExistingOnDate} and \textit{endsExistingOnDate}.
Facts are assigned a time point if they are instantaneous events or a time span if they have an extended duration
with known beginning and end.
This approach allows the system to deduce the temporal scope of entities and events, providing a comprehensive temporal annotation across the dataset.

For example, entities such as people are assigned birth and death dates, while artifacts and groups have creation and potential destruction dates.
In cases where the data is incomplete or not applicable (e.g.,~abstract concepts or mythological entities), no temporal data is assigned,
adhering to a conservative approach.
This temporal framework is essential for enabling time-based queries and analyses,
such as determining the lifespan of individuals or the duration of historical events.

YAGO 4.5~\cite{yago45} builds upon the temporal framework of previous versions by enhancing its capacity to handle temporal information more flexibly and in more detail.
The integration of temporal data into YAGO 4.5 utilizes the RDF-star model, allowing for more intricate annotations of facts.
This temporal tagging is crucial for representing the evolving nature of the KG, allowing for dynamic queries that reflect changes over time.
All YAGO versions are accessible only as archives, as the live version is not publicly available.

\subsubsection{Wikidata}

Wikidata~\cite{DBLP:conf/www/Vrandecic12} utilizes a system of qualifiers to provide additional context to statements, including temporal information.
Qualifiers in Wikidata are versatile and can specify details such as the period during which a statement is valid.
For instance, in the following example, Douglas Adams's education at St John's College is annotated with start and end dates,
providing a temporal dimension to his academic history:

\begin{verbatim}
wd:Q42 wdt:P69 wd:Q3918 .
wd:Q42 p:P69 _:statement .
_:statement ps:P69 wd:Q3918 .
_:statement pq:P580 "1971-10-01T00:00:00Z"^^xsd:dateTime .
_:statement pq:P582 "1974-06-01T00:00:00Z"^^xsd:dateTime .
\end{verbatim}

In this data snippet, `\textit{wd:Q42}' represents Douglas Adams, `\textit{wd:Q3918}' represents St John's College, `\textit{p:P69}' and
`\textit{ps:P69}' denote the property for educational institutions attended,
and `\textit{pq:P580}' and `\textit{pq:P582}' provide the start and end times of the education period, respectively.

Qualifiers in Wikidata allow for highly detailed annotations of statements, including dates, locations, and other contextual information.
This feature supports complex data representation but requires sophisticated data parsing and querying mechanisms to retrieve and
interpret the full context of the data.

Wikidata's approach to versioning involves maintaining a complete history of all edits, thus allowing for comprehensive tracking of changes over time.
This version history is essential for verifying data provenance and understanding the evolution of knowledge within the graph.
Unlike other KGs, Wikidata's version history is openly accessible, providing a transparent view of the data's editorial process.

Furthermore, Wikidata is a DKG, continuously updated by a large community of contributors.
This live updating system ensures that Wikidata remains current, reflecting the latest information and corrections as they become available.
The dynamic nature of Wikidata, combined with its detailed versioning and temporal annotations,
supports both real-time applications and historical research, making it a versatile tool for a wide range of uses.

\subsubsection{EventGraph}

EventKG~\cite{DBLP:conf/esws/GottschalkD18,DBLP:conf/esws/GottschalkD18a}, a multilingual event-centric TKG,
focuses heavily on capturing and representing temporal relations and events.
The system models events using a canonical representation that incorporates both the start and end times of events,
leveraging a variety of data sources, including Wikidata, DBpedia, and YAGO.
Temporal information in EventKG is linked to entities and events, allowing for detailed historical analysis and event tracking.
The model uses properties such as \textit{sem:hasBeginTimeStamp} and \textit{sem:hasEndTimeStamp} to define the temporal span of events and
entities' involvement in those events.

Additionally, EventKG employs the Simple Event Model (SEM) as its foundational schema, which it extends to represent temporal relations better.
This includes modeling not only event-entity relationships but also complex temporal relations between multiple entities or events,
such as sequences of events and their hierarchical relationships.
This comprehensive temporal modeling enables sophisticated queries and analyses concerning the temporal dynamics of historical and
contemporary events.

Using named graphs and quadruples, EventKG also addresses versioning indirectly by maintaining a provenance framework that tracks the sources and
version history of the data it integrates.
Provenance information is critical in EventKG, as it helps verify the accuracy and credibility of the temporal data.
Each piece of information, including temporal annotations, is associated with metadata that details its origin,
the specific version of the source data, and the extraction date.
\section{Dynamic Knowledge Graph Completion}\label{sec:kgcompletion}


In the existing literature~\cite{wang2023survey}, TKGC methods are generally classified into two categories: interpolation-based and extrapolation-based. Interpolation-based methods aim to predict missing knowledge by leveraging existing quadruplets. In contrast, extrapolation-based methods are designed for continuous TKGs, enabling prediction of future events by learning embeddings from previous or historical snapshots of entities and relations. 
This work treats interpolation methods as TKGC for TKGs, while extrapolation-based methods are applied to dynamic KGs, as defined in Section~\ref{sec:prelimiaries}.

\subsection{Temporal Knowledge Graph Completion (TKGC)} \label{subsec:tkgc} 

Temporal Knowledge Graphs (TKGs) often contain millions or billions of quadruplets, but they are typically incomplete for several reasons. First, extracting information from unstructured text sources can be error-prone, resulting in incomplete data. Second, capturing or integrating all available information, particularly from diverse or complex sources, is challenging. Third, the sources themselves may lack comprehensiveness or be biased, leading to selective inclusion of information and omission of other relevant facts. Lastly, the dynamic nature of information, with knowledge continuously evolving, contributes to these gaps.

The KGC task -- commonly known as link prediction (LP) -- aims to predict missing links by utilizing existing information. 
Various techniques have been developed for this task. However, a significant limitation of these methods is their difficulty in capturing the temporal dynamics of facts. They are typically designed for static knowledge graphs (KGs) that assume facts do not change over time. Therefore, these methods are ineffective when applied to TKGs, as they overlook the crucial temporal information inherent in TKGs. 
Temporal KGC (TKGC) methods have emerged to enhance LP accuracy by addressing the limitations of traditional KGC methods. TKGC methods improve upon these by incorporating timestamps of facts into the learning process in addition to the facts themselves.  For instance, with TKGC, it is possible to infer the fact \texttt{(Gerhard Schröder, succeeded, Angela Merkel, 2005)} using the existing quadruplets \texttt{(Angela Merkel, Chancelor of, Germany, [2005 to 2021])} and \texttt{(Gerhard Schröder, Chancelor of, Germany, [1998 - 2005])}.  

TKGC aims to predict possible links between two entities at a specific time. This can be accomplished in different ways: i) \textit{tail prediction} - given the head and relation at a certain time, predicting the tail entity ($<h,r,?>$), ii) \textit{head prediction} - given the tail and relation at a particular time, predicting the head entity ($<?,r,t>$), iii) \textit{relation prediction} - given the head and tail at a certain time, predicting the relation ($<h,?,t>$). Inspired by the survey in~\cite{cai2022temporal}, TKGC methods could be categorized into timestamps-dependent-based TKGC, timestamps-specific functions-based TKGC, and deep learning-based TKGC methods. A summary of these TKGC methods is provided in Table~\ref{tab:tkgc-summary}. Timestamps dependent-based TKGC methods 
such as TuckERTNT~\cite{shao2022tucker}, TTransE~\cite{leblay2018deriving}, ST-TransE~\cite{ni2020specific}, T-SimplE~\cite{lin2020tensor}, Canonical Polyadic (CP)~\cite{he2023improving}, TKGFrame~\cite{zhang2020tkgframe}, TBDRI~\cite{yu2023tbdri} associate timestamps to corresponding entities and relations to capture their evolution without directly manipulating the timestamps. 

Timestamps-specific Functions-based TKGC methods use specialized functions, such as diachronic embedding, Gaussian, and transformation function, to learn embeddings for timestamps. Transformation functions-based TKGC methods include BoxTE~\cite{messner2022temporal}, SPLIME~\cite{radstok2021leveraging}, TARGCN~\cite{ding2112simple}, TASTER~\cite{wang2023temporal}, Time-LowFER~\cite{dikeoulias2022temporal}, Goel et al.~\cite{goel2020diachronic}, and DEGAT~\cite{wang2022dynamic}. Complex embedding functions-based TKGC methods include  ChronoR~\cite{sadeghian2021chronor}, TComplEx and TNTComplEx~\cite{lacroix2020tensor}, TeRo~\cite{xu2020tero}, TGeomE~\cite{xu2022geometric}, TeLM~\cite{xu2021temporal}, RotateQVS~\cite{chen2022rotateqvs}, ST-NewDE~\cite{nayyeri2022dihedron}, BiQCap~\cite{zhang2023biqcap},HA-TKGE~\cite{zhang2022hierarchy}, STKE~\cite{wang2022stke}, and HTKE~\cite{he2022hyperplane}. Non-linear embedding functions-based TKGC methods iclude DyERNIE~\cite{han2020dyernie}, ATiSE~\cite{xu2020temporal}, HERCULES~\cite{montella2021hyperbolic}, and TKGC-AGP~\cite{zhang2022temporal}.

Deep learning-based TKGC methods capture the evolution of entities and relations by encoding timestamps using deep learning algorithms. These methods can be further grouped into i) Timestamps-Specific Space such as HyTE~\cite{dasgupta2018hyte}, HTKE~\cite{he2022hyperplane}, TRHyTE~\cite{yuan2022trhyte}, BTHyTE~\cite{liu2021temporal}, ToKEi~\cite{leblay2020towards}, SANe~\cite{li2022each}, and QDN~\cite{wang2023qdn}, ii) Long Short-Term Memory (LSTM)-based TKGC methods such as TA-TransE and TA-DistMult~\cite{garcia2018learning}, TDG2E~\cite{tang2020timespan}, Ma et al~\cite{ma2021learning}, CTRIEJ~\cite{li2023embedding}, and TeCre~\cite{ma2023tecre}, and iii) Temporal constraint-based TKGC methods such as Chekol et al.~\cite{chekol2017marrying}, Kgedl~\cite{wang2019novel}, T-GAP~\cite{jung2021learning}, TempCaps~\cite{fu2022tempcaps}, RoAN~\cite{bai2023roan}, and TAL-TKGC~\cite{nie2023temporal}. 



\begin{table}[]
    \centering
    \caption{Comparative Analysis of the TKGC methods presented in Section~\ref{subsec:tkgc}. ILP, CS, and SKGC stand for Integer Linear Programming, Coordinate System, and Static Knowledge Graph Completion, respectively.}
    \begin{tabular}{c|p{2cm}|p{4.4cm}}
    \hline
       \textbf{Technique } & \textbf{TKGC Methods} & \textbf{Remarks}\\
        \hline
       Translation & TTransE~\cite{leblay2018deriving}, ST-TransE~\cite{ni2020specific}, BoxTE~\cite{messner2022temporal} & TTransE and ST-TransE struggle with handling temporally evolving facts.  All are SKGC methods. \\
       Tensor Decomp. & TuckERTNT~\cite{shao2022tucker}, T-SimplE~\cite{lin2020tensor}, Time-LowFER~\cite{dikeoulias2022temporal}, QDN~\cite{wang2023qdn} & Time-LowFER ignores the rich contextual information in the graph structure. All except QDN are SKGC methods\\
       ILP & TKGFrame~\cite{zhang2020tkgframe} & TKGFrame can not handle complex scenarios like path queries.\\
       Manifold & HERCULES~\cite{montella2021hyperbolic}, DyERNIE~\cite{han2020dyernie} & Both are SKGC methods. HERCULES does not show significant
differences over benchmarks due to its limited use of temporal information.\\
       Block Decomp. & TBDRI~\cite{yu2023tbdri} & TBDRI is an SKGC method.\\
       GCN & TARGCN~\cite{ding2112simple} & TARGCN is an SKGC method.\\
       GAT & DEGAT~\cite{wang2022dynamic} & \\
       Transformation & SPLIME~\cite{radstok2021leveraging} & SPLIME is an SKGC method.\\
       Sparse Matrix &  TASTER~\cite{wang2023temporal} & TASTER is an SKGC method.\\
       Gaussian &  ATiSE~\cite{xu2020temporal} & \\
       Gaussian/Markov &  TKGC-AGP~\cite{zhang2022temporal} & \\
       Polar CS &  HA-TKGE~\cite{zhang2022hierarchy}, HTKE~\cite{he2022hyperplane} & Both are SKGC methods.\\
       Spherical CS &  STKE~\cite{wang2022stke} & \\
       Complex Space &  TNTComplEx~\cite{lacroix2020tensor}, TeRo~\cite{xu2020tero}, ChronoR~\cite{sadeghian2021chronor} & All are SKGC methods.\\
       Quaternion Space & TeLM~\cite{xu2021temporal}, RotateQVS~\cite{chen2022rotateqvs}, TGeomE~\cite{xu2022geometric} & All are SKGC methods. TeLM ignores the rich contextual
information in the graph structure.\\
       Dihedron Algebra & ST-NewDE~\cite{nayyeri2022dihedron} & ST-NewDE is an SKGC method.\\
       Biquaternions/Manifold &  BiQCap~\cite{zhang2023biqcap} & BiQCap is an SKGC method.\\
       Time-Encoding & TA-DistMult~\cite{garcia2018learning}, ToKEi~\cite{leblay2020towards} & Both are SKGC methods.\\
       Time-Encoding/LSTM &  TeCre~\cite{ma2023tecre} & TeCre is an SKGC method.\\
       Temporal-Hyperp & HyTE~\cite{dasgupta2018hyte}, BTHyTE~\cite{liu2021temporal} &  Both are SKGC methods.\\
       Temporal-Hyperp/GRU & TRHyTE~\cite{yuan2022trhyte} & TRHyTE is an SKGC method.\\
       Multi-semantic Space & SANe~\cite{li2022each} & \\
       GRU & TDG2E~\cite{tang2020timespan}\\
       GRU/Negative Sampling & CTRIEJ~\cite{li2023embedding} & CTRIEJ is an SKGC method.\\
       Path Reasoning & Kgedl~\cite{wang2019novel} & \\
       GNN & T-GAP~\cite{jung2021learning} & \\
       Capsule Network & TempCaps~\cite{fu2022tempcaps} & \\
       Attention & RoAN~\cite{bai2023roan}, TAL-TKGC~\cite{nie2023temporal} & Both are SKGC methods.\\
       \hline
    \end{tabular}
    \label{tab:tkgc-summary}
\end{table}
\paragraph{\textbf{Training procedure}:} Given a TKG $G$ and a set of facts $Q$ in $G$, 
a scoring function $g(q)$ is defined to assign a factual score for a fact or a true quadruplet $q \in Q$. A negative sampling strategy~\cite{yang2020understanding} is employed to create a set of negative samples, i.e.,~factually incorrect quadruplets $Q'$, to enhance the expressiveness of learned representations for the entities and relations in $G$.  

\paragraph{\textbf{Loss function}:} A loss function $L$ aims at maximizing $g(q)$ for all $q \in Q$ and minimizing $g(q')$ for their negative samples $q' \in Q$. The following are the commonly used loss functions by TKGC approaches.
\begin{itemize}
    \item \textbf{Margin-based ranking loss (MRL)}~\cite{bordes2013translating} enforces that the confidence in the corrupted quadruplet is lower than in the true quadruplet by a certain margin. MRL is computed as 
    \begin{equation}
        L_{MRL} = \sum_{q \in Q}{\big[\lambda + g(q) - \sum_{q'\in Q'}{g(q')}\big]_{+}},
    \end{equation}
    where $[x]_+ = max(x, 0)$ and $\lambda > 0$ is a margin hyperparameter.
    \item \textbf{Cross-entropy loss (CEL)} \cite{li2021search} also aims at obtaining a large gap between true quadruplets and the negative samples but without enforcing a fixed margin for all facts. 
        \begin{equation}
        L_{CEL} = \sum_{q \in Q}\frac{exp(g(q))}{\sum_{q'\in Q'}{exp(g(q'))}}
    \end{equation}
    
    \item \textbf{Binary cross-entropy loss (BCEL)}~\cite{liu2019context} emphasizes the score of individual true quadruplets and negative samples as follows: 
     \begin{equation}
        L_{BCEL} = \sum_{q \in Q \cup Q'} yg(q) + (1-y)g(q'),
    \end{equation}
   where $y = 1$ if $q \in Q$ and $y = 0$ otherwise. 
\end{itemize}
\paragraph{\textbf{Evaluation techniques:}} Typically, evaluating TKGC methods involves performing both head and tail predictions for each test quadruple $q$. For head prediction, the head entity in $q$ is replaced with every possible entity in the TKG. Similarly, the tail entity is replaced with every possible entity for tail prediction. The scores generated by the scoring function for these modified quadruples are then ranked. Commonly used accuracy metrics for this evaluation include Mean Rank (MR), Mean Reciprocal Rank (MRR), and Hits@k.

\paragraph{\textbf{Benchmark datasets:}} The commonly used benchmark datasets for TKGC can be categorized into those used for interpolation methods, such as ICEWS14~\cite{garcia2018learning}, ICEWS05-15~\cite{garcia2018learning}, GDELT~\cite{leetaru2013gdelt}, YAGO11k~\cite{dasgupta2018hyte}, YAGO15k~\cite{garcia2018learning}, and Wikidata12k~\cite{dasgupta2018hyte}, and those used for extrapolation methods, including ICEWS14~\cite{garcia2018learning}, ICEWS18~\cite{jin2019recurrent}, GDELT~\cite{leetaru2013gdelt}, WIKI~\cite{leblay2018deriving}, and YAGO~\cite{mahdisoltani2013yago3}.

\subsection{Non-Temporal Dynamic KG Completion} 

Traditional methods for generating KG embeddings do not consider the evolving nature of a KG where new entities and relations are constantly being added to the KG. This requires training the embeddings over the new KG from scratch even if the changes are minimal which can lead to increased computational costs. Various attempts have been made to deal with this challenge. 
One of the first attempts to address this issue is online learning, which learns incrementally as new information arrives. 
However, one of the drawbacks of these methods of KG embedding is that they do not consider the problem of catastrophic forgetting. To mitigate this issue, continual or lifelong learning~\cite{pami/WangZSZ24} methods were proposed, alleviating the problem of catastrophic forgetting in various tasks. In catastrophic forgetting, adaptation to a new distribution generally results in a largely reduced ability to capture the old ones.

Continual KG Embedding methods (CKGE) has recently received growing attention which perform fine-tuning with only new knowledge, leading to reduced training costs. These methods effectively alleviate the problem of catastrophic forgetting while learning the representations of the newly added knowledge. Continual Learning-based methods further target two kinds of solutions. First, the full-parameter fine-tuning paradigm which memorizes old knowledge by replaying a core old dataset or introducing additional regularization constraints. Although this paradigm effectively mitigates catastrophic forgetting, it significantly increases training costs, especially when handling large-scale KGs. Second, it adopts the incremental-parameter fine-tuning paradigm, with only a few parameters to learn emerging knowledge. This strategy may still lead to an increase in parameters and training time. 

Recently, low-rank adapters such as LoRA have enabled efficient parameter fine-tuning and are used to reduce the training time in Large Language Models (LLMs). One of the very recent studies, FastKGE, introduces incremental low-rank adaptation mechanisms, IncLoRAs, to reduce training costs for continual learning for KG embeddings. The rest of the section discusses and compares these methods in detail. 

\subsubsection{Online Learning Based Methods}
\paragraph{puTransE~\cite{TayLH17}.} 

The current translational models (KG embedding models using translation-based scoring functions) for static KG embeddings have three significant shortcomings. 
First, translational models are highly sensitive to hyperparameters such as margin and learning rate~\cite{AliBHVGSFTL22}. Second, for one triple, there is only one representation if the translation principle is followed, leading to low precision due to the congestion of entities and relations in vector space. Last but not least, new unseen entities or relations are not handled. One of the first methods to deal with the drawbacks mentioned above is Parallel Universe TransE (puTransE) -- an extension of TransE -- which learns multiple embeddings. 

puTransE follows three steps: (i) triple selection (structurally and semantically aware triple selection), (ii) generation of random configurations, and (iii) learning embeddings.

 \begin{description}
    \item[Triple Selection] Triple Selection includes semantically and structurally aware triple selection. To select semantically relevant entities, the puTransE samples a relation $r \in R$ and generates a set of entities $\{e_1, e_2, \dots, e_n\} \in E$ of all entities containing $r$ as either an outgoing or incoming edge. For structurally aware triple selection, the authors adopt the bidirectional random walk model using the entities selected as starting nodes. puTransE creates multiple embedding spaces each time new entities or relations arrive, and for each embedding, it defines a triple constraint both in count and diversity.   
    \item[Generate Random Configurations] Instead of defining a global configuration, puTransE generates different hyperparameter configurations for each embedding space. It randomly generates the value of the original TransE hyperparameters, i.e.,~margin and learning rate but also the values of triple constraint and number of training epochs.
    \item[Learning Embeddings] puTransE uses a margin-based loss function. 
\end{description}

puTransE, however, learns embeddings from the local parts of the KG, avoiding the retraining of the whole embedding space, which leads to the loss of global structural information in the learned embedding space. Additionally, since puTransE is an increment over TransE, it uses the scoring function of TransE, which does not work well for 1-to-N, N-to-1, and N-to-N relations. 

\paragraph{DKGE \cite{kbs/WuKYQW22}} learns joint embedding for each entity and relation by considering their contextual information leading to an improvement over puTransE which only considers local information. The context of an entity $e_1$ is the one-hop neighborhood subgraph of the entity represented as $sg(e_1) \in E$. The context of a relation $r_1$ is represented by the relation path $p_1 = (r_1, r_2)$. The first step is to encode the contextual information of entity and relation as vector representations, which is then combined with the entity and relation embeddings (referred to as knowledge embeddings). After that, a scoring function and a loss function based on translation operations for parameter training are defined. 

The algorithm uses Attentive GCN (AGCN) Model to effectively take into account the neighborhood information (modeled by GCN) and only use the information that may be important (modeled by the attention mechanism). Given $x$ where $x \in E or R$ and its context, i.e., a subgraph with $n$ vertices $\{v_i\}^n_{i =1}$, DKGE builds the adjacency matrix $A \in \mathbb{R}^{n \times n}$ and initializes the vertex feature matrix $H^{(0)} \in \mathbb{R}^{n\times d^0}$ where $d^0$ is the number of the initialized features for each vertex. Each row in $H^{(0)}$ is denoted as $v_i$. If $x$ is an entity, then $v_i$ is an entity and ${\bf v_i}$ is its contextual element embedding. When $x$ is a relation, if $v_i$ is a relation path consisting of two relations, then ${\bf v_i}$ is the sum of the contextual element embeddings of these two relations. 

The final contextual subgraph embedding ${\bf sg}(x)$ of the subgraph $sg(x)$ is given by a weighted sum of the vectors of all vertices $\{v_i\}^n_{i=1}$ as follows:

\begin{equation}\label{eq:scoring_dkge}
{\bf sg}(x) = \sum^{n}_{i=n} \alpha_{i(x)} {\bf v_i} 
\end{equation}

\noindent The scoring function is given as follows:

\begin{equation}
    f(h,r,t) = \parallel \textbf{h*} + \textbf{r*} - \textbf{t*} \parallel_{l_1}
\end{equation}
where \textbf{h*}, \textbf{r*} and \textbf{t*} are computed by Equation~\ref{eq:scoring_dkge}, and $\parallel \odot \parallel_{l_1}$ denotes the ${l_1}$ norm. A margin-based loss function is utilized for training. During online learning, at time step $T+1$, the knowledge embeddings and contextual element embeddings for new entities (relations) are randomly initialized according to a uniform distribution. For existing entities (relations), their knowledge embeddings and contextual element embeddings are carried over from the embedding results at time step $T$. These methods, however, do not consider literal information along with the structural information of a KG.

DKGC-JSTD~\cite{ica3pp/XieWWZF20} proposes a DKG completion model that jointly learns the structural and textual information of entities and relations based on a deep RNN. This model learns the embedding of an entity’s name and parts of its text description to connect unseen entities to KGs. Deep memory network and association matching mechanism are used to extract semantic feature information and establish relevance between entity and relations from entity text-description. It then uses Recurrent Neural Networks to model the dependency between topology-structure and text description.

\subsubsection{Continual Learning Based Methods}

Various methods based on continual learning have been proposed. In this chapter, we discuss some of the fundamental algorithms.

\paragraph{Lifelong Knowledge Graph Embedding (LKGE)~\cite{CuiWSLJHH23}} learns the embedding of an entity or relation based on its masked first-order subgraph given as follows: 

\begin{equation}
     \overline{x}_i  = MAE(\bigcup^{i}_{j=1}N_j(x))
\end{equation}

where $x \in E~or~R$, and $N_j \subseteq D_j$ denotes the involved facts of $x$ in the $j$-th snapshot and D represents the training set. $MAE()$ is an encoder that represents the input subgraph. The objective of the KG encoder is to align the entity or relation embedding with the reconstructed representation as follows:

\begin{equation}
    L_{MAE} = \sum_{e \in E_i} \parallel e_i - \overline{e}_i \parallel^2_2 +  \sum_r \in R_i  \parallel r_i - \overline{r}_i \parallel^2_2
\end{equation}

Graph Convolution Networks (GCN) and Transformer are two common encoders. If they are updated for lifelong learning, the changed model parameters will affect the embedding generation of all old entities, including the involved old entities in new facts. This can lead to catastrophic forgetting of the previous snapshots. For this reason, the entity and relation embedding transition functions are used as encoders by LKGE, which do not introduce additional parameters and use TransE. TransE is leveraged to train the embeddings from the new data and update the knowledge transfer for each snapshot.

\begin{equation}
    L_{new} = \sum_{(h,r,t)\in D_i} max (0,\gamma + f(h,r,t) - f(h',r,t'))
\end{equation}

where $\gamma$ is the margin. $(h',r,t')$ is the embedding of a negative fact. The subject or object is randomly replaced with a random entity $e' \in E_i$ for each positive fact. The embeddings of unseen entities $\textbf{e}$ and relations $\textbf{r}$ are randomly initialized. To avoid catastrophic forgetting, regularization methods constrain the updates of the parameters. Accordingly, the loss function of regularization methods is given as:

\begin{equation}
    L_{old} = \sum_{e\in E_{i-1}} w(e)\parallel e_i - e_{i-1} \parallel^2_2 + \sum_r \in R_{i-1} w(r) \parallel r_i - r_{i-1} \parallel^2_2 
\end{equation}

where $w(x)$ is the regularization weight for $x$. The regularization weight is only computed once per snapshot. The overall learning objective is as follows:

\begin{equation}
L = L_{new} + \alpha L_{old} + \beta L_{MAE}
\end{equation}

where $\alpha, \beta$ are hyperparameters for balancing the objectives.


One of the recent approaches uses low-rank adaptation for learning dynamic KG embeddings. Incremental Low-Rank Adaptation (IncLoRA)~\cite{liu2024fastcontinualknowledgegraph} operated in three stages. During the first graph layering stage, new entities and relations are divided into several layers based on the distance from the old graph and node degrees. 
The second stage involves IncLoRA learning, in which incremental LoRAs with adaptive rank allocation represent the embeddings of entities and relations in each layer. 
In the final stage, link prediction is performed, which composes all new LoRAs into a LoRA group and concat all LoRA groups and initial embeddings for inference. 

Several previous studies have proposed using different mechanisms to learn the representations of the initial state of KGs and then incrementally learning the representations upon the arrival of new entities and relations. Some continual learning based algorithms use translational embeddings~\cite{access/SongP18}, progressive neural networks~\cite{ral/DarunaGSC21}, incremental knowledge distillation~\cite{aaai/LiuK0SGLJL24}, and graph attention networks~\cite{emnlp/KouLLLZZ20}.

\section{Dynamic Entity Alignment}\label{sec:entity_alignment}

This section provides an overview of the temporal entity alignment methods between two KGs. In addition to TKG alignment, a few methods have been proposed considering new data while performing entity alignment. 

\subsection{Temporal Entity Alignment Methods}

Most of the methods for temporal entity alignment use GCNs or GNNs to learn the representations of the entities alongside the temporal attention mechanism. In the following, we briefly overview the methods proposed so far.

In~\cite{corr/abs-2203-02150}, the authors exploit both relation and temporal information for entity alignment by first creating a reverse link, i.e.,~introducing the inverse of the relations to handle the beginning and end of the relation. The method utilizes a time-aware attention mechanism with GCN to assign different weights to entities according to their relation and temporal information. On the other hand, TKG Entity Alignment via Representation Learning (Tem-EA)~\cite{dasfaa/SongBLZ22} incorporates temporal information with the help of Recurrent Neural Networks to learn temporal sequence representations. GCN and translation-based embedding models are used for learning representations over structural and attribute information for computing entity similarity separately on the two KGs to be aligned. These representations are combined using linear weighting. The concept of nearest-neighbor matching is performed to find the most similar entity pair based on the distance matrix.

A recent study~\cite{coling/CaiMMYZL22} proposes a simple temporal information matching mechanism for entity alignment between TKGs. Most of the methods proposed prior to this use a time-aware attention mechanism to incorporate relational and temporal information into entity embeddings. This study assumes that learning the representations from the temporal information is unnecessary since a temporal matching mechanism, in addition to the GNN-based model, achieves better results. In~\cite{kbs/LiSW0ZZ23}, the authors perform entity alignment for TKGs via adaptive graph networks. More specifically, a time-aware graph attention network model is used as an encoder to aggregate the features and temporal relationships of neighboring nodes. DualMatch~\cite{www/LiuW00G23} is an unsupervised method that fuses the relational and temporal information for EA by encoding temporal and relational information into embeddings using a dual-encoder and combining both the information and transforming it into alignment using a novel graph-matching-based decoder called GM-Decoder.

\subsection{Temporal and Evolving Entity Alignment Methods}

\paragraph{Temporal Relational Entity Alignment (TREA)~\cite{www/XuSX022}} learns alignment-oriented TKG embeddings and represents new emerging entities. The first step is to map entities, relations, and timestamps into an embedding space, and the initial feature of each entity is represented by fusing the embeddings of its connected relations and timestamps as well as its neighboring entities. A Graph Neural Network (GNN) is employed to capture intra-graph information, and a temporal relational attention mechanism is utilized to integrate relation and time features of links between nodes. Finally, a margin-based full multi-class log-loss is used for efficient training, and a sequential time regularizer is used to model unobserved timestamps.

\paragraph{Incremental Temporal Entity Alignment (ITEA)~\cite{DBLP:conf/icdm/LiCLZL23}} targets the problem of temporality as well as evolving KGs. It uses a combination of knowledge distillation with the Graph Attention Network (GAT) and the GCN acting as the teacher and student models, respectively. The proposed model transfers knowledge from a complex model, the teacher, whose output (in terms of probabilities) is used to train a simpler model, the student. The teacher Model within knowledge distillation provides instructional guidance to the student model during its training phase.   

As a first stage, entity embeddings are generated from the entity labels obtained using GloVe. For preserving structural information in the teacher model, a masked attention mechanism over the neighboring entities of the entity $e_i$. Time-aware structure embedding is obtained using a multi-head attention-based GNN. This allows the teacher model to not only learn the node features and time-aware structural information but also adjust their contributions adaptively based on the requirements of the specific task. The student model uses an importance-sampling strategy to select a small subset of nodes or neighbors in each layer, and the model is trained on mini-batches of nodes.

Entity alignment module aligns the newly incoming entities with the existing ones, and these alignments are achieved through string matching, structural similarity, etc. The margin-based logistic loss function measures the dissimilarity between the embeddings of aligned entities.
The output from the teacher model is collected by passing the new data through the teacher model to compute its output probabilities, $P_t$. Then, the output of the student model is collected by passing the same data through the student model to compute its output probabilities, $P_s$. The total loss $L$ can be written as:

$$L = \alpha_1 \times L_{standard} + \alpha_2 \times L_{distillation}$$

where $\alpha_1$ and $\alpha_2$ are weights that determine the relative importance of the standard loss and the distillation. Grid search is used to define possible values for $\alpha_1$ and $\alpha_2$. The standard loss compares the student's outputs $P_s$ with the true labels $y$. $L_{standard}$ for a classification task is typically the cross-entropy loss given as:

$$L_{standard} = - \sum y \times log(P_s) $$

where the sum is over all classes, and $y$ is a one-hot encoded vector of the true labels. The distillation loss that compares the student’s outputs $P_s$ with the teacher’s outputs $P_t$. $L_{distillation}$ is the Kullback-Leibler divergence between the teacher’s outputs and the student’s outputs and is given as:

$$L_{distillation} = \sum P_t \times log(\frac{P_t}{P_s})$$

\section{Discussion \& Future Directions}\label{sec:discussion}

This chapter formally defines various types of DKGs and explores how this knowledge can be represented within KGs. It then delves into different representation learning techniques for both non-temporal and temporal DKGs, focusing on tasks related to KG completion. Additionally, it covers methods for aligning temporal and dynamic KGs. The studies discussed in this chapter primarily rely on the triple structure or treat the KG as a graph by incorporating the contextual information of entities and relationships while largely overlooking schematic or ontological information~\cite{corr/abs-2308-00081}. Furthermore, the adaptability of these algorithms to domain-specific and real-world applications is limited, especially as KGs can expand to contain billions or even trillions of triples. Many of these algorithms also fail to utilize the literal information within KGs. Only one of the recent studies has attempted to use Large Language Models (LLMs) as background information for TKGC~\cite{corr/abs-2401-06072} based on few-shot learning and concluded that LLMs do not bring significant increases in performance.  However, the study misses the analysis of the results in the sense of how many predictions were hallucinations or overgenerations of LLMs for the specific task.

\bibliographystyle{vancouver}
\bibliography{references}

\end{document}